\documentclass{article}

 \usepackage[final,nonatbib]{nips_2017}

\usepackage[utf8]{inputenc} % allow utf-8 input
\usepackage[T1]{fontenc}    % use 8-bit T1 fonts
\usepackage{hyperref}       % hyperlinks
\usepackage{url}            % simple URL typesetting
\usepackage{booktabs}       % professional-quality tables
\usepackage{amsfonts, amsmath}   % blackboard math symbols
\usepackage{nicefrac}       % compact symbols for 1/2, etc.
\usepackage{microtype}      % microtypography
\usepackage{algorithm}
\usepackage{algorithmic}

\usepackage[pdftex]{graphicx}
\usepackage[tight,footnotesize]{subfigure}

\title{Accelerated Primal-Dual Policy Optimization for Safe Reinforcement Learning}

\author{
  Qingkai Liang \\
  MIT\\
  \texttt{qingkai@mit.edu} \\
   \And
   Fanyu Que \\
   Boston College \\
   \texttt{quef@bc.edu} \\
   \And
   Eytan Modiano \\
   MIT \\
   \texttt{modiano@mit.edu} \\
}

\begin{document}

\maketitle

\begin{abstract}
Constrained Markov Decision Process (CMDP) is a natural framework for reinforcement learning tasks with safety constraints, where agents learn a policy that maximizes the long-term reward while satisfying the constraints on the long-term cost. A canonical approach for solving CMDPs is the primal-dual method which updates parameters in primal and  dual spaces in turn. Existing methods for CMDPs only use on-policy data for dual updates, which results in sample inefficiency and slow convergence. In this paper, we propose a policy search method for CMDPs called Accelerated Primal-Dual Optimization (APDO), which incorporates an off-policy trained dual variable in the dual update procedure while updating the policy in primal space with on-policy likelihood ratio gradient. Experimental results on a simulated robot locomotion task show that APDO achieves better sample efficiency and faster convergence than state-of-the-art approaches for CMDPs.
\end{abstract}

\section{Introduction}
In reinforcement learning (RL), agents learn to act by trial and error in an unknown environment. The majority of RL algorithms allow agents to freely explore the environment and exploit any actions that might improve the reward. However,  actions that lead to high rewards usually come with high risks. In a safety-critical environment, it is important to enforce safety in the RL algorithm, and a natural way to enforce safety is to incorporate constraints. A standard formulation for RL with safety constraints is the constrained Markov Decision Process (CMDP) framework \cite{CMDP}, where the agents need to maximize the long-term reward while satisfying the constraints on the long-term cost.
Applications of CMDPs include windmill control \cite{wind} where we need to maximize the average reward (e.g., generated power) while bounding the long-term wear-and-tear cost on critical components (e.g., wind turbine). Another important example is communication network control where we need to maximize network utility while bounding the long-term arrival rate below the long-term service rate in order to maintain network stability (Chapter 1.1 in \cite{CMDP}).

While optimal policies for finite CMDPs with known models can be obtained by linear programming \cite{LP1}, it cannot scale to high-dimensional continuous control tasks  due to curse of dimensionality.  Recently, there have been RL algorithms that work for high-dimensional CMDPs based on advances in policy search algorithms \cite{TRPO, A3C}. In particular, two constrained policy search algorithms enjoy state-of-the-art performance for CMDPs:  Primal-Dual Optimization (PDO) \cite{PDO} and Constrained Policy Optimization (CPO)  \cite{CPO}. PDO is based on  Lagrangian relaxation and updates parameters in primal and  dual spaces in turn. Specifically, the primal policy update uses the policy gradient descent while the dual variable update uses the dual gradient ascent. By comparison, CPO differs from PDO in the dual update procedure, where the dual variable is obtained from scratch by solving a carefully-designed optimization problem in each iteration, in order to enforce safety constraints throughout training. Besides PDO and CPO, there exist other methods for solving CMDPs \cite{uchibe2007constrained, ammar2015safe, held2017probabilistically}, but these approaches are usually computationally intensive or only apply to some specific CMDP models and domains.

A notable feature of existing constrained  policy search approaches (e.g., PDO and CPO) is that they only use \emph{on-policy} samples\footnote{On-policy samples refer to those generated by the currently-used policy while off-policy samples are generated by other unknown policies.}, which ensures that the information used for dual updates is unbiased and leads to stable performance improvement. However, such an on-policy dual update is sample-inefficient since historical samples are discarded. Moreover, due to the on-policy nature, dual updates are incremental and suffer from slow convergence since a (potentially large) batch of on-policy samples have to be obtained before a dual update can be made. 

In this paper, we propose a policy search method for CMDPs called Accelerated Primal-Dual Optimization (APDO), which incorporates an off-policy trained dual variable in the dual update procedure while updating the policy in primal space with on-policy likelihood ratio gradient. Specifically, APDO is similar to PDO except that we perform a one-time adjustment for the dual variable with a nearly optimal dual variable trained with off-policy data after a certain number of iterations. Such a one-time adjustment process incurs negligible amortized overhead in the long term but greatly improves the sample efficiency and the convergence rate over exisiting methods. We demonstrate the effectiveness of APDO on a simulated robot locomotion task where the agent must satisfy constraints motivated by safety. The experimental results show that APDO achieves better sample efficiency and faster convergence than state-of-the-art approaches for CMDPs (e.g., PDO and CPO).

Another line of work considers merging the on-policy and off-policy policy gradient updates to improve sample efficiency. Examples of these approaches include Q-Prop \cite{Q-prop}, IPG \cite{IPG}, etc. These approaches are designed for unconstrained MDPs and can be applied to the primal policy update. In contrast, APDO leverages off-policy samples for dual updates and is complementary to these efforts on merging on-policy and off-policy policy gradients. %In fact, the dual update procedures in APDO can be combined with any of these sophisticated policy update strategies to further improve sample efficiency. 
\section{Constrained Markov Decision Process}
A Markov Decision Process (MDP) is represented by a tuple, $(\mathcal{S},\mathcal{A},R,P,p_0)$, where $\mathcal{S}$ is the set of states, $\mathcal{A}$ is the set of actions, $R:\mathcal{S}\times\mathcal{A}\times\mathcal{S}\mapsto \mathbb{R}$ is the reward function, $P:\mathcal{S}\times\mathcal{A}\times\mathcal{S}\mapsto [0,1]$ is the transition probability function (where $P(s'|s,a)$  is the transition probability from state $s$ to state $s'$ given action $a$), and $p_0:\mathcal{S}\mapsto [0,1]$ is the initial state distribution. A stationary policy $\pi:\mathcal{S}\mapsto \mathcal{P}(\mathcal{A})$ corresponds to a mapping from states to a probability distribution over actions. Specifically, $\pi(a|s)$ is the probability of selecting action $a$ in state $s$. The set of all stationary policies is denoted by $\Pi$. In this paper, we search policy within a parametrized stationary policy class $\Pi_{\theta}\subset \Pi$ (e.g., a neural network policy class with weight $\theta$). We may write a policy $\pi$ as $\pi(\theta)$ to emphasize its dependence on the parameter $\theta$. The long-term discounted reward under policy $\pi$ is denoted as $R(\pi)=\mathbb{E}_{\tau\sim \pi}[\sum_{t=0}^{\infty} \gamma^t R(s_t,a_t, s_{t+1})]$, where $\gamma\in[0,1)$ is the discount factor, $\tau=(s_0,a_0,s_1,a_1,\cdots)$ denotes a trajectory, and $\tau\sim \pi$ means that the distribution over trajectories is determined by policy $\pi$, i.e., $s_0\sim p_0,a_t\sim \pi(\cdot|s_t),s_{t+1\sim P(\cdot|s_t,a_t)}$.  

A constrained Markov Decision Process (CDMP) is an MDP augmented with constraints on long-term discounted costs. Specifically, we augment the ordinary MDP with $m$ cost functions $C_1,\cdots,C_m$, where each cost function $C_i:\mathcal{S}\times\mathcal{A}\times\mathcal{S}\mapsto \mathbb{R}$ is a mapping from transition tuples to costs. The long-term discounted cost under policy $\pi$ is similarly defined as $C_i(\pi)=\mathbb{E}_{\tau\sim \pi}[\sum_{t=0}^{\infty} \gamma^t C_i(s_t,a_t, s_{t+1})]$, and the corresponding limit is  $d_i$. In CMDP, we aim to select a policy $\pi$ that maximizes the long-term reward $R(\pi)$ while satisfying the constraints on the long-term costs $C_i(\pi)\le d_i,~\forall i\in [m]$, i.e.,
\begin{equation}\label{eq:cmdp}
\begin{split}
\pi^*=&\arg\max_{\pi\in \Pi_{\theta}}~~R(\pi)\\
&\text{s.t.}~~C_i(\pi)\le d_i,~\forall i\in [m].
\end{split}
\end{equation}
\section{Algorithm}
To solve CMDPs, we employ the Lagrangian relaxation procedure (Chapster 3 in \cite{lagrange}). Specifically, the Lagrangian function for the CMDP problem \eqref{eq:cmdp} is
\begin{equation}
\mathcal{L}(\pi,\lambda)=R(\pi)-\sum_i\lambda_i \Big(C_i(\pi)-d_i\Big),
\end{equation}
where $\lambda=(\lambda_1,\cdots,\lambda_m)$ is the Lagrangian multiplier. Then the constrained problem \eqref{eq:cmdp} can be converted to the following unconstrained problem:
\begin{equation}\label{eq:minimax}
(\pi^*,\lambda^*)=\arg\min_{\lambda \ge 0}\max_{\pi\in\Pi_{\theta}} \mathcal{L}(\pi,\lambda).
\end{equation}
%In addition, under mild technical conditions, it was shown in \cite{CMDP} that for CMDPs we have
%\begin{equation}
%\min_{\lambda \ge 0}\max_{\pi\in\Pi_{\theta}} \mathcal{L}(\pi,\lambda) = \max_{\pi\in\Pi_{\theta}} \min_{\lambda \ge 0}\mathcal{L}(\pi,\lambda).
%\end{equation}
To solve the unconstrained minimax problem \eqref{eq:minimax}, a canonical approach is to use the iterative primal-dual method where in each iteration we update the primal policy $\pi$ and the dual variable $\lambda$ in turn. The primal-dual update procedures at iteration $k$ are as follows:

\vspace{1mm}

\hspace{0mm} $\bullet$ \quad Fix $\lambda=\lambda^{(k)}$ and perform policy gradient update: 
$
\theta_{k+1}=\theta_k + \alpha_k \nabla_{\theta} (\mathcal{L}(\pi(\theta),\lambda^{(k)}))|_{\theta=\theta_k},
$
where $\alpha_k$ is the step size. The policy gradient could be on-policy likelihood ratio policy gradient (e.g., REINFORCE \cite{REINFORCE} and TRPO \cite{TRPO}) or off-policy deterministic policy gradient (e.g., DDPG \cite{DDPG}).

\vspace{1mm}

\hspace{0mm} $\bullet$ \quad Fix $\pi=\pi_k$ and perform dual update $\lambda^{(k+1)}=f_k(\lambda^{(k)}, \pi_k)$. Existing methods for CMDPs, such as PDO and CPO, differ in the choice of the dual update procedure $f_k(\cdot)$. For example, PDO uses the simple dual gradient ascent
$
\lambda_i^{(k+1)}=[\lambda_i^{(k)}+\beta_k(C_i(\pi_k)-d_i)]^+,
$
where $\beta_k$ is the step size and  $[x]^+=\max\{0,x\}$ is the projection onto the dual space $\lambda\ge 0$. By comparison, CPO derives the dual variable $\lambda^{(k+1)}$ by solving an optimization problem from scratch in order to enforce the constraints in every iteration.

\vspace{1mm}

However, the dual update procedures used in existing methods (e.g., PDO and CPO) are incremental and only use on-policy samples, resulting sample inefficiency and slow convergence to the  optimal primal-dual solution $(\lambda^*, \pi^*)$. In this paper, we propose to incorporate an off-policy trained dual variable in the dual update procedure in order to improve sample efficiency and speed up the search for the optimal dual variable $\lambda^*$. The algorithm is called Accelerated Primal-Dual Optimization (APDO) and is described in  Algorithm \ref{alg:apdo}. APDO is similar to PDO  where in most iterations the dual variable is updated according to the simple dual gradient ascent (step \ref{step:dual-gradient}), but the key innovation of APDO is that there is a one-time dual adjustment with an off-policy trained dual variable $\lambda^{\text{OFF}}$ after $K^{\text{adj}}$ iterations (steps \ref{step:adjust1}-\ref{step:adjust2}). The off-policy trained $\lambda^{\text{OFF}}$ is obtained by running an off-policy  algorithm for CMDPs with the historical data stored in the replay buffer. We provide a primal-dual version of the DDPG algorithm in the supplementary material for training $\lambda^{\text{OFF}}$.  Although the off-policy trained dual variable $\lambda^{\text{OFF}}$ could be biased, it provides a nearly optimal point for further fine tuning of the dual variable using new on-policy data.

The improvement of sample efficiency in APDO is due to the fact that off-policy training can repeatedly exploit historical data while on-policy update only uses each sample once; the acceleration effect of APDO is due to the fact that off-policy training directly solves for the optimal dual variable offline, thus avoiding the slow on-policy dual update as in the existing approaches where only one dual update can be taken after a large batch of samples are obtained. 

Note that the adjustment epoch $K^{\text{adj}}$ is an important parameter in APDO. Using a small $K^{\text{adj}}$ avoids slow incremental dual update early, but the dual estimate $\lambda^{\text{OFF}}$ could be highly biased and inaccurate due to insufficient amount of data. On the other hand, using a larger $K^{\text{adj}}$ provides a more accurate dual estimate at the expense of delayed adjustment. %One future work is to provide theoretical guidance on the setting of $K^{\text{adj}}$. 

%Note that the primal policy update in APDO can use any on-policy likelihood ratio policy gradient method (e.g., REINFORCE \cite{REINFORCE} and TRPO \cite{TRPO}), or even more sophisticated policy optimization methods, such as Q-Prop \cite{Q-prop} and IPG \cite{IPG}, that merge on-policy and off-policy policy gradients.

\begin{algorithm}\label{alg:apdo}
\caption{Accelerated Primal-Dual Policy Optimization (APDO)}\label{alg:tracking}
\begin{algorithmic}[1]
\STATE Initialize policy $\pi_0\in\Pi_{\theta}$, replay buffer $\mathcal{R}=\emptyset$
\FOR{$k=0,1,2\cdots,$}
\STATE Sample a set of trajectories $\mathcal{D}_k$ under the current policy $\pi_k=\pi(\theta_k)$ (containing $T$ samples)
\STATE Add the sampled data $\{(s_t,a_t,r_t,c_t,s_{t+1})\}_{t=0}^T$ to the replay buffer $\mathcal{R}$
\STATE Update the primal policy with any on-policy likelihood ratio gradient method (e.g., TRPO) using the sampled on-policy trajectories $\mathcal{D}_k$ and the current dual variable $\lambda^{(k)}$
\STATE Update the dual variable with dual gradient ascent: $\lambda_i^{(k+1)}=\Big[\lambda_i^{(k)}+\beta_k\Big(C_i(\pi_k)-d_i\Big)\Big]^+,\forall i$ \label{step:dual-gradient}
\IF {$k=K^{\text{adj}}$}\label{step:adjust1}
\STATE Compute the off-policy trained $\lambda^{\text{OFF}}$ with the replay buffer $\mathcal{R}$ (e.g., using the primal-dual DDPG in the supplementary material)
\STATE Set $\lambda^{(k+1)}=\lambda^{\text{OFF}}$
\ENDIF\label{step:adjust2}
\ENDFOR
\end{algorithmic}
\end{algorithm}
 
\section{Experiments}
We evaluate APDO against two state-of-the-art algorithms for solving CMDPs (i.e., CPO and PDO) on a simple point-gather control task in MuJoCo \cite{mujoco} with an additional safety constraint as used in \cite{CPO}. All experiments are implemented in rllab \cite{rllab}.  The detailed task description and experiment parameters are provided in the supplementary material. In particular, for APDO we set the adjustment epoch $K^{\text{adj}}=5$, and additional experimental results regarding the effect of $K^{\text{adj}}$ are also given in the supplementary material. 

\begin{figure*}[ht!]
%\centering
\subfigure[Average Return]{\label{fig:point-gather-return}\includegraphics[height=33mm]{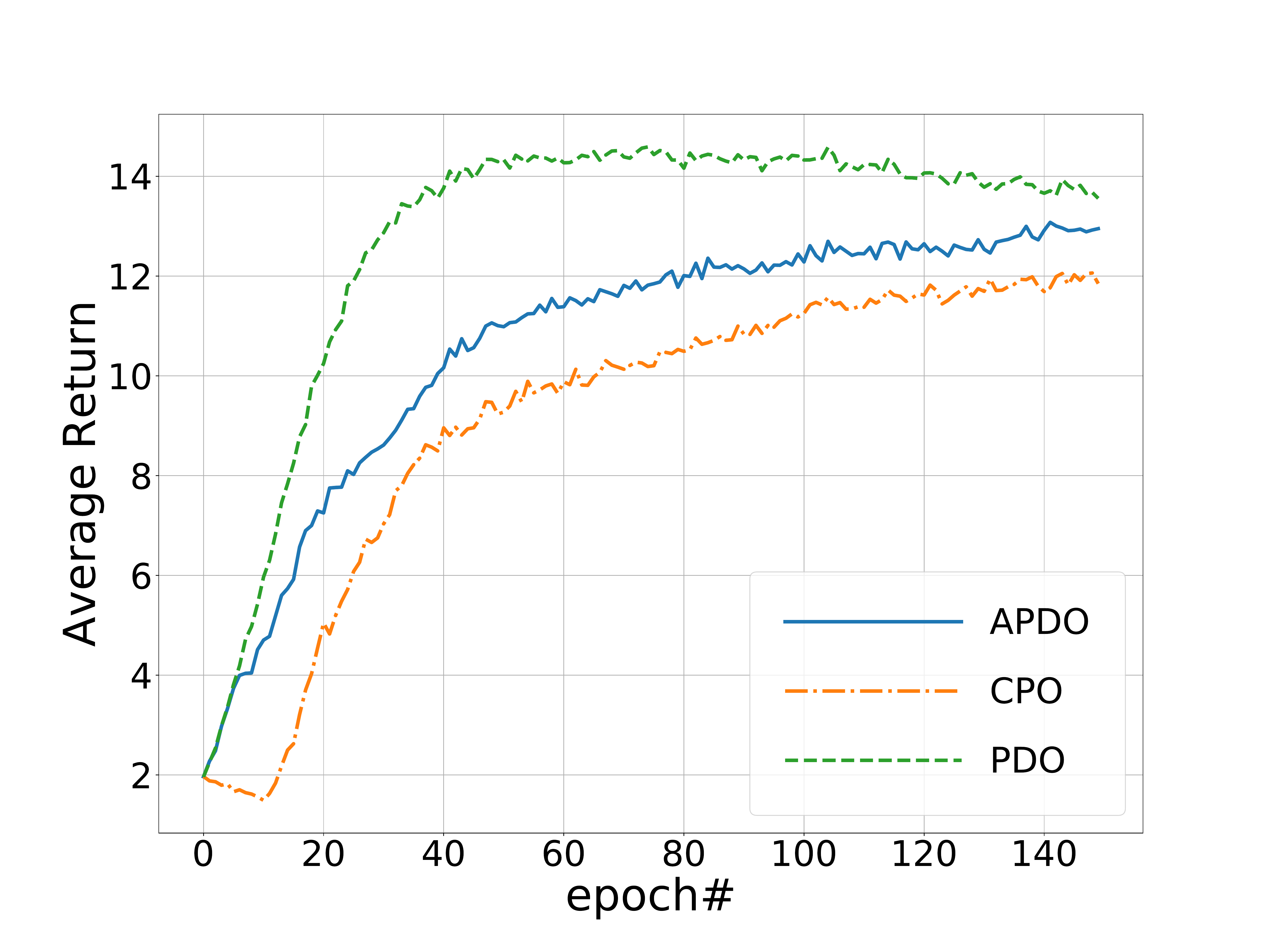}}
\subfigure[Average Cost (limit is 0.2)]{\label{fig:point-gather-cost}\includegraphics[height=33mm]{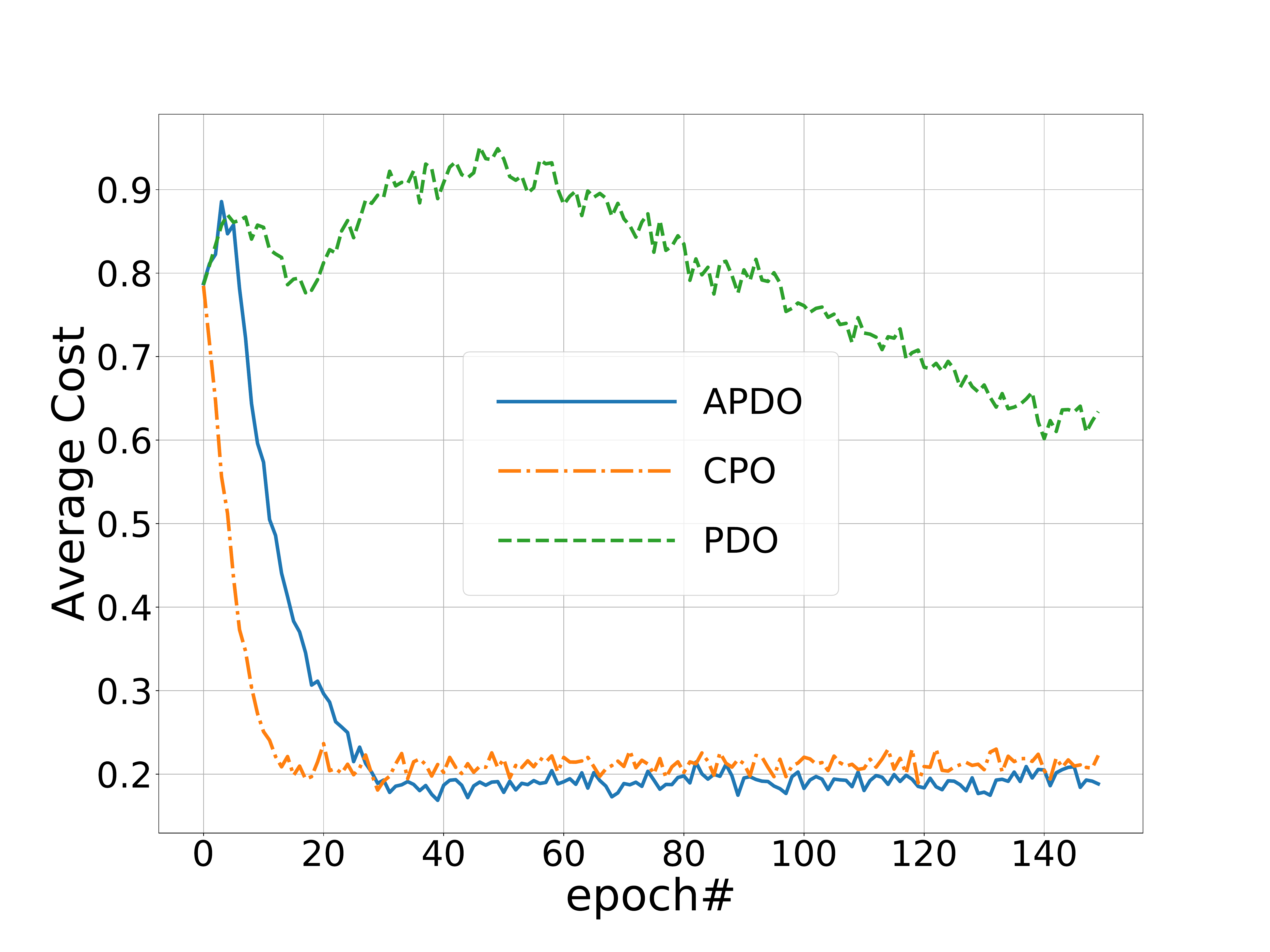}}
\subfigure[Dual Variable]{\label{fig:point-gather-dual}\includegraphics[height=33mm]{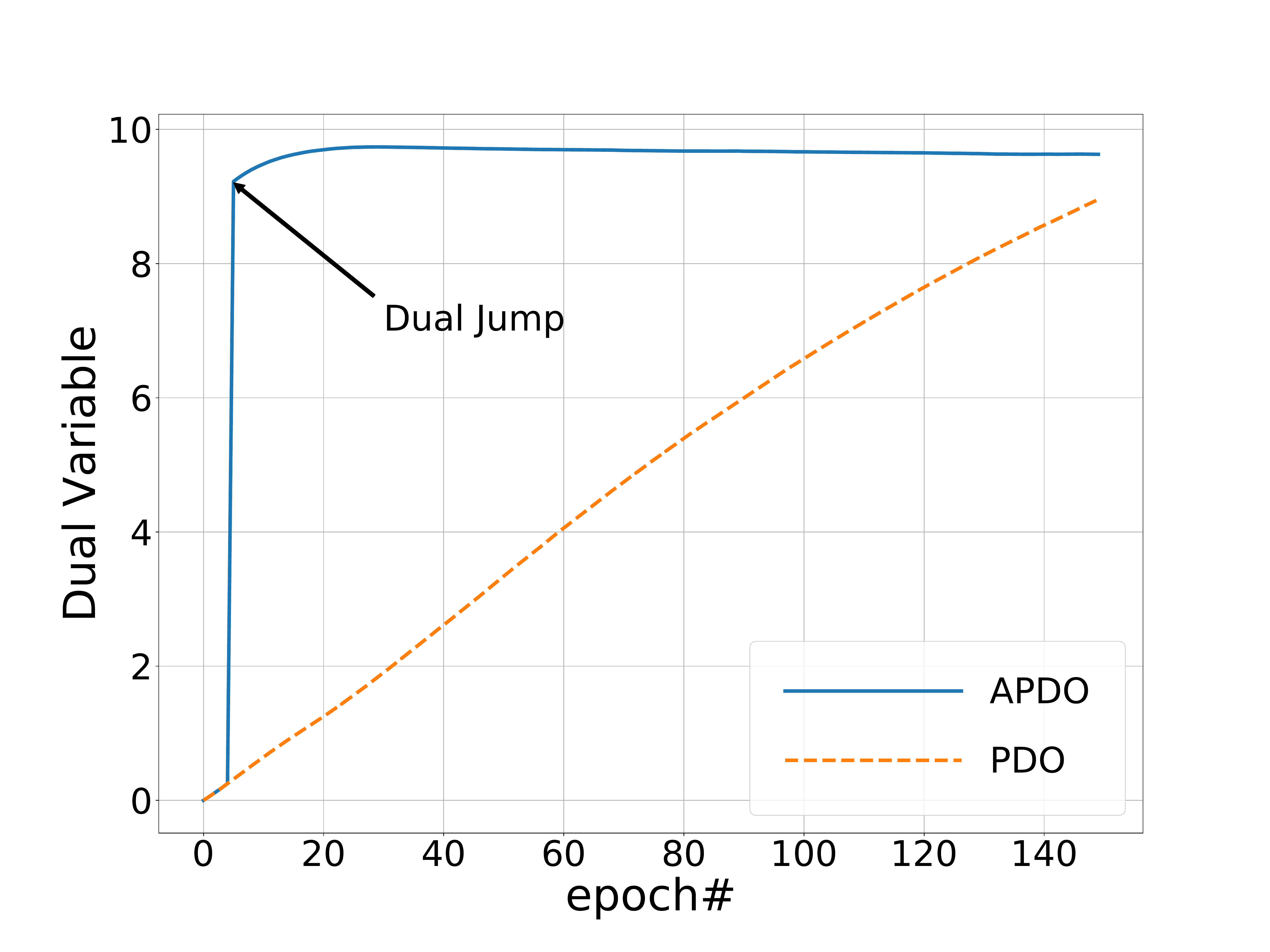}}
\caption{Performance comparison among APDO, PDO and CPO.}
\label{fig:point-gather-performance}\vspace{-3mm}
\end{figure*}
Figure \ref{fig:point-gather-performance} shows the learning curves for APDO, CPO and PDO under cost constraints. It can be observed from Fugure \ref{fig:point-gather-cost} that APDO enforced constraints successfully to the limit value as approximately same speed as CPO did. More importantly, APDO generally outperforms CPO on reward performance without compromising constraint stabilization, thus achieving better sample efficiency. For example, CPO takes 90 epochs to achieve an average reward of 11 while satisfying the safety constraint. By comparison, APDO only takes 45 epochs to achieve the same point, which corresponds to 2x improvement in sample efficiency over CPO in this task. In addition, PDO fails to enforce the safety constraint during the first 150 epochs due to its slow convergence. Using a larger step size may help speed up the convergence but in this case PDO will over-correct in response to constraint violations and behave too conservatively. We provide additional discussions on the choice of stepsize for PDO and APDO in the supplementary material.

Figure \ref{fig:point-gather-dual} illustrates the learning trajectory of the dual variable under PDO and APDO (note that the dual variable for CPO is not illustrated since CPO has a sophisticated recovery scheme to enforce constraints, where the dual variable may not be easily obtained). We find that APDO converges to the optimal dual variable $\lambda^*$ significantly faster than PDO. In particular, there is a ``jump" of the dual variable after several epochs in APDO, due to the dual adjustment with the off-policy trained $\lambda^{\text{OFF}}$. By comparison, PDO has to adjust its dual variable incrementally with on-policy data.

\section{Future Work}
Since the adjustment epoch is an important parameter in APDO, one important future work is to provide theoretical guidance on the setting of $K^{\text{adj}}$.  It is also very interesting (yet challenging) to provide theoretical justifications about the acceleration effects of APDO. Moreover, as we observed in the experiments, the training trajectory generated by APDO strives for the best tradeoff between improving rewards and enforcing cost constraints. One future work is to incorporate a safety parameter that controls the degree of safety awareness. By tuning the parameter, the RL algorithm should be able to make both risk-averse actions (which enforce safety constraints as soon as possible) and risk-neutral actions (which gives priority to improving rewards). 

\section*{Acknowledgment}
This work was supported by NSF Grant CNS-1524317 and by DARPA I2O and Raytheon BBN Technologies under Contract No. HROO l l-l 5-C-0097. The authors would also like to acknowledge Chengtao Li who provided valuable feedback on this work.

\newpage

\bibliographystyle{IEEEtran}
\bibliography{Bib/Qingkai-APDO}

\begin{thebibliography}{10}
\providecommand{\url}[1]{#1}
\csname url@samestyle\endcsname
\providecommand{\newblock}{\relax}
\providecommand{\bibinfo}[2]{#2}
\providecommand{\BIBentrySTDinterwordspacing}{\spaceskip=0pt\relax}
\providecommand{\BIBentryALTinterwordstretchfactor}{4}
\providecommand{\BIBentryALTinterwordspacing}{\spaceskip=\fontdimen2\font plus
\BIBentryALTinterwordstretchfactor\fontdimen3\font minus
  \fontdimen4\font\relax}
\providecommand{\BIBforeignlanguage}[2]{{%
\expandafter\ifx\csname l@#1\endcsname\relax
\typeout{** WARNING: IEEEtran.bst: No hyphenation pattern has been}%
\typeout{** loaded for the language `#1'. Using the pattern for}%
\typeout{** the default language instead.}%
\else
\language=\csname l@#1\endcsname
\fi
#2}}
\providecommand{\BIBdecl}{\relax}
\BIBdecl

\bibitem{safe}
S.~Shalev-Shwartz, S.~Shammah, and A.~Shashua, ``Safe, multi-agent,
  reinforcement learning for autonomous driving,'' \emph{arXiv preprint
  arXiv:1610.03295}, 2016.

\bibitem{CMDP}
E.~Altman, \emph{Constrained Markov decision processes}.\hskip 1em plus 0.5em
  minus 0.4em\relax CRC Press, 1999, vol.~7.

\bibitem{wind}
P.~Tavner, J.~Xiang, and F.~Spinato, ``Reliability analysis for wind
  turbines,'' \emph{Wind Energy}, vol.~10, no.~1, pp. 1--18, 2007.

\bibitem{LP1}
E.~A. Feinberg and A.~Shwartz, ``Constrained dynamic programming with two
  discount factors: Applications and an algorithm,'' \emph{IEEE Transactions on
  Automatic Control}, vol.~44, no.~3, pp. 628--631, 1999.

\bibitem{TRPO}
J.~Schulman, S.~Levine, P.~Abbeel, M.~Jordan, and P.~Moritz, ``Trust region
  policy optimization,'' in \emph{Proceedings of the 32nd International
  Conference on Machine Learning (ICML-15)}, 2015, pp. 1889--1897.

\bibitem{A3C}
V.~Mnih, A.~P. Badia, M.~Mirza, A.~Graves, T.~Lillicrap, T.~Harley, D.~Silver,
  and K.~Kavukcuoglu, ``Asynchronous methods for deep reinforcement learning,''
  in \emph{International Conference on Machine Learning}, 2016, pp. 1928--1937.

\bibitem{PDO}
Y.~Chow, M.~Ghavamzadeh, L.~Janson, and M.~Pavone, ``Risk-constrained
  reinforcement learning with percentile risk criteria,'' \emph{arXiv preprint
  arXiv:1512.01629}, 2015.

\bibitem{CPO}
J.~Achiam, D.~Held, A.~Tamar, and P.~Abbeel, ``Constrained policy
  optimization,'' in \emph{Proceedings of the 34nd International Conference on
  Machine Learning (ICML-17)}, 2017.

\bibitem{uchibe2007constrained}
E.~Uchibe and K.~Doya, ``Constrained reinforcement learning from intrinsic and
  extrinsic rewards,'' in \emph{Development and Learning, 2007. ICDL 2007. IEEE
  6th International Conference on}.\hskip 1em plus 0.5em minus 0.4em\relax
  IEEE, 2007, pp. 163--168.

\bibitem{ammar2015safe}
H.~B. Ammar, R.~Tutunov, and E.~Eaton, ``Safe policy search for lifelong
  reinforcement learning with sublinear regret,'' in \emph{Proceedings of the
  32nd International Conference on Machine Learning (ICML-15)}, 2015, pp.
  2361--2369.

\bibitem{held2017probabilistically}
D.~Held, Z.~McCarthy, M.~Zhang, F.~Shentu, and P.~Abbeel, ``Probabilistically
  safe policy transfer,'' \emph{arXiv preprint arXiv:1705.05394}, 2017.

\bibitem{Q-prop}
S.~Gu, T.~Lillicrap, Z.~Ghahramani, R.~E. Turner, and S.~Levine, ``Q-prop:
  Sample-efficient policy gradient with an off-policy critic,''
  \emph{International Conference on Learning Representations (ICLR-17)}, 2017.

\bibitem{IPG}
S.~Gu, T.~Lillicrap, Z.~Ghahramani, R.~E. Turner, B.~Sch{\"o}lkopf, and
  S.~Levine, ``Interpolated policy gradient: Merging on-policy and off-policy
  gradient estimation for deep reinforcement learning,'' \emph{arXiv preprint
  arXiv:1706.00387}, 2017.

\bibitem{lagrange}
D.~P. Bertsekas, \emph{Nonlinear programming}.\hskip 1em plus 0.5em minus
  0.4em\relax Athena scientific Belmont, 1999.

\bibitem{REINFORCE}
R.~J. Williams, ``Simple statistical gradient-following algorithms for
  connectionist reinforcement learning,'' \emph{Machine learning}, vol.~8, no.
  3-4, pp. 229--256, 1992.

\bibitem{DDPG}
T.~P. Lillicrap, J.~J. Hunt, A.~Pritzel, N.~Heess, T.~Erez, Y.~Tassa,
  D.~Silver, and D.~Wierstra, ``Continuous control with deep reinforcement
  learning,'' \emph{International Conference on Learning Representations
  (ICLR-16)}, 2016.

\bibitem{mujoco}
E.~Todorov, T.~Erez, and Y.~Tassa, ``Mujoco: A physics engine for model-based
  control,'' in \emph{Intelligent Robots and Systems (IROS), 2012 IEEE/RSJ
  International Conference on}.\hskip 1em plus 0.5em minus 0.4em\relax IEEE,
  2012, pp. 5026--5033.

\bibitem{rllab}
Y.~Duan, X.~Chen, R.~Houthooft, J.~Schulman, and P.~Abbeel, ``Benchmarking deep
  reinforcement learning for continuous control,'' in \emph{International
  Conference on Machine Learning}, 2016, pp. 1329--1338.

\bibitem{GAE}
J.~Schulman, P.~Moritz, S.~Levine, M.~Jordan, and P.~Abbeel, ``High-dimensional
  continuous control using generalized advantage estimation,'' \emph{arXiv
  preprint arXiv:1506.02438}, 2015.

\bibitem{adam}
D.~Kingma and J.~Ba, ``Adam: A method for stochastic optimization,''
  \emph{arXiv preprint arXiv:1412.6980}, 2014.

\end{thebibliography}

\newpage

\section*{Supplementary Materials}

\appendix

\section{Primal-Dual DDPG for CMDPs}\label{ap:ddpg}
In this appendix, we provide a primal-dual version of the DDPG algorithm for solving CMDPs. The primal policy update and the dual variable update in this algorithm only use the off-policy data stored in the replay buffer, which can be used to fit  $\lambda^{\text{OFF}}$ for our APDO algorithm. For simplicity, we only present the algorithm for CMDPs with a single constraint, and the multiple-constraint case can be easily obtained. In the primal-dual DDPG algorithm, we have the following neural networks.
\begin{itemize}
\item{Reward critic Q-network $Q_{R}(s,a|\theta_R^Q)$ and reward target Q-network $Q'_R(s,a|\theta_R^{Q'})$}
\item{Cost critic Q-network $Q_{C}(s,a|\theta_C^Q)$ and cost target Q-network $Q'_C(s,a|\theta_C^{Q'})$}
\item{Actor policy network $\mu(s|\theta^\mu)$ and actor target Q-network $\mu'(s|\theta^{\mu'})$}
\end{itemize}
The target networks are used to slowly track the learned networks.
\begin{algorithm}\label{alg:ddpg}
\caption{Primal-Dual DDPG}
\begin{algorithmic}[1]
\STATE Randomly initialize reward critic Q-network $Q_{R}(s,a|\theta_R^Q)$, cost critic Q-network $Q_{C}(s,a|\theta_C^Q)$ and actor  network $\mu(s|\theta^\mu)$ 
\STATE Initialize target networks: $\theta_R^{Q'}\leftarrow \theta_R^{Q}$, $\theta_C^{Q'}\leftarrow \theta_C^Q$, $\theta^{\mu'}\leftarrow \theta^{\mu}$
\STATE Initialize replay buffer $\mathcal{R}$ and dual variable $\lambda$
\FOR{episode $k=0,1,\cdots,$}
\STATE Initialize a random process $\mathcal{N}$ for action exploration
\STATE Receive initial state $s_0\sim p_0$
\FOR{$t=1,\cdots,T$}
\STATE Select action $a_t=\mu(s_t|\theta^\mu)+\mathcal{N}_t$
\STATE Execute action $a_t$ and observe $r_t, c_t, s_{t+1}$
\STATE Store transition $(s_t,a_t,r_t,c_t,s_{t+1})$ in the replay buffer $\mathcal{R}$
\STATE Sample a random batch of $N$ transitions $\{(s_i,a_i,r_i,c_i,s_{i+1})\}_{i=1}^N$ from the replay buffer $\mathcal{R}$
\STATE Set $y_i=r_i+\gamma Q_R'(s_{i+1},\mu'(s_{i+1}|\theta^{\mu'})|\theta_R^{Q'}))$, $z_i=c_i+\gamma Q_C'(s_{i+1},\mu'(s_{i+1}|\theta^{\mu'})|\theta_C^{Q'}))$
\STATE Update reward critic by minimizing $L_R=\frac{1}{N}\sum_i (y_i-Q_R(s_i,a_i|\theta_R^{Q}))^2$ and update cost critic by minimizing $L_C=\frac{1}{N}\sum_i (z_i-Q_C(s_i,a_i|\theta_C^{Q}))^2$
\STATE Update the actor policy using the sampled policy gradient
\[
\nabla_{\theta^{\mu}} \mathcal{L}(\theta^\mu, \lambda)=\frac{1}{N}\sum_i \nabla_{\theta^\mu}\Big(Q_R(s,\mu(s|\theta^\mu)|\theta_R^Q)-\lambda Q_C(s,\mu(s|\theta^\mu)|\theta_C^Q)\Big)\Big|_{s=s_i}
\]
\STATE Update dual variable using the sampled dual gradient
\[
\nabla_{\lambda} \mathcal{L}(\theta^\mu, \lambda) = \frac{1}{N}\sum_i \Big[Q_C(s_i,\mu(s_i|\theta^{\mu}))-d\Big]
\]
\STATE Update target networks:
\[
\begin{split}
&\theta_R^{Q'}\leftarrow \tau \theta_R^Q+(1-\tau)\theta_R^{Q'}\\
&\theta_C^{Q'}\leftarrow \tau \theta_C^Q+(1-\tau)\theta_C^{Q'}\\
&\theta^{\mu'}\leftarrow \tau \theta^\mu+(1-\tau)\theta^{\mu'}\\
\end{split}
\]
\ENDFOR
\ENDFOR
\end{algorithmic}
\end{algorithm}

\section{Experiment Details}\label{ap:exp}

\noindent \textbf{Task description.} Specifically, a point mass receives a reward of 10 for collecting an apple, and a cost of 1 for collecting a bomb.  The agent is constrained to incur no more than 0.2 cost in the long term. Two apples and eight bombs spawn on the map at the start of each episode.

\noindent \textbf{Parameters for primal policy update.} For all experiments, we use neural network policies with two hidden layers of sizes $(64,32)$ with tanh non-linearity, and all of the schemes (PDO, CPO, APDO) use TRPO to update the primal policy, with a batch size 50000 and a KL-divergence step size of 0.01. The discount factor is 0.995 and the rollout length is 15. We use GAE-$\lambda$ \cite{GAE} for estimating the regular advantages with $\lambda^{\text{GAE}}=0.95$.

\noindent \textbf{Parameters for dual variable update.}  As for dual updates, PDO and APDO both use dual gradient ascent. Note that the step size for dual gradient ascent is important in PDO: if it is set to be too small, the dual variable won't update quickly enough to meaningfully
enforce the constraint; if it is too high, the algorithm will over-correct in response to constraint violations and behave too conservatively \cite{CPO} As a result, picking a proper step size is critical and difficult in PDO. We experiment with different step sizes $\{0.01, 0.05, 0.1, 0.5\}$ and find that 0.1 works best for PDO, and the reported results of PDO are also under the step size 0.1. By comparison, selecting step size in APDO is much easier since the one-time off-policy dual adjustment directly boosts the dual variable to a "nearly optimal" point and we only need to choose a relatively small step size in order to do fine-tuning after the adjustment. For the reported experimental results, we also set the step size to be 0.1 for APDO for the fairness of comparison. As for CPO, we adopt the same set of parameters as in original CPO paper \cite{CPO} (specially, the parameters used in the point-gather task).

\noindent \textbf{Parameters for training $\lambda^{\text{OFF}}$.} We use primal-dual DDPG to train $\lambda^{\text{OFF}}$. The reward critic network  $Q_{R}(s,a|\theta_R^Q)$ and cost critic network $Q_{C}(s,a|\theta_C^Q)$ ) is parametrized by a neural network with two hidden layers of sizes $(100,100)$ with tanh nonlinearity, respectively. The actor policy network $\mu(s|\theta^\mu)$ is represented by a neural network with two hidden layers of sizes $(64,32)$ with tanh nonlinearity. The learning rates for the reward/cost critic Q-network and the actor policy network are all $10^{-3}$ and these networks are updated with Adam \cite{adam}. The update for the dual variable in primal-dual DDPG employs simple dual gradient ascent and the step size for updating the dual variable in the primal-dual DDPG is set to be $10^{-2}$. The mini-batch size is $N=64$. We also use a soft target networks with $\tau=0.001$. The off-policy training is executed for $5\times 10^5$ primal-dual iterations. Since off-policy algorithms like DDPG are usually unstable, we set $\lambda^{\text{OFF}}$ to be the average of all historical dual variables throughout the off-policy training trajectory. The max replay buffer size is $10^6$.

\begin{figure*}[ht!]
%\centering
\subfigure[Average Return]{\label{fig:k1}\includegraphics[height=33mm]{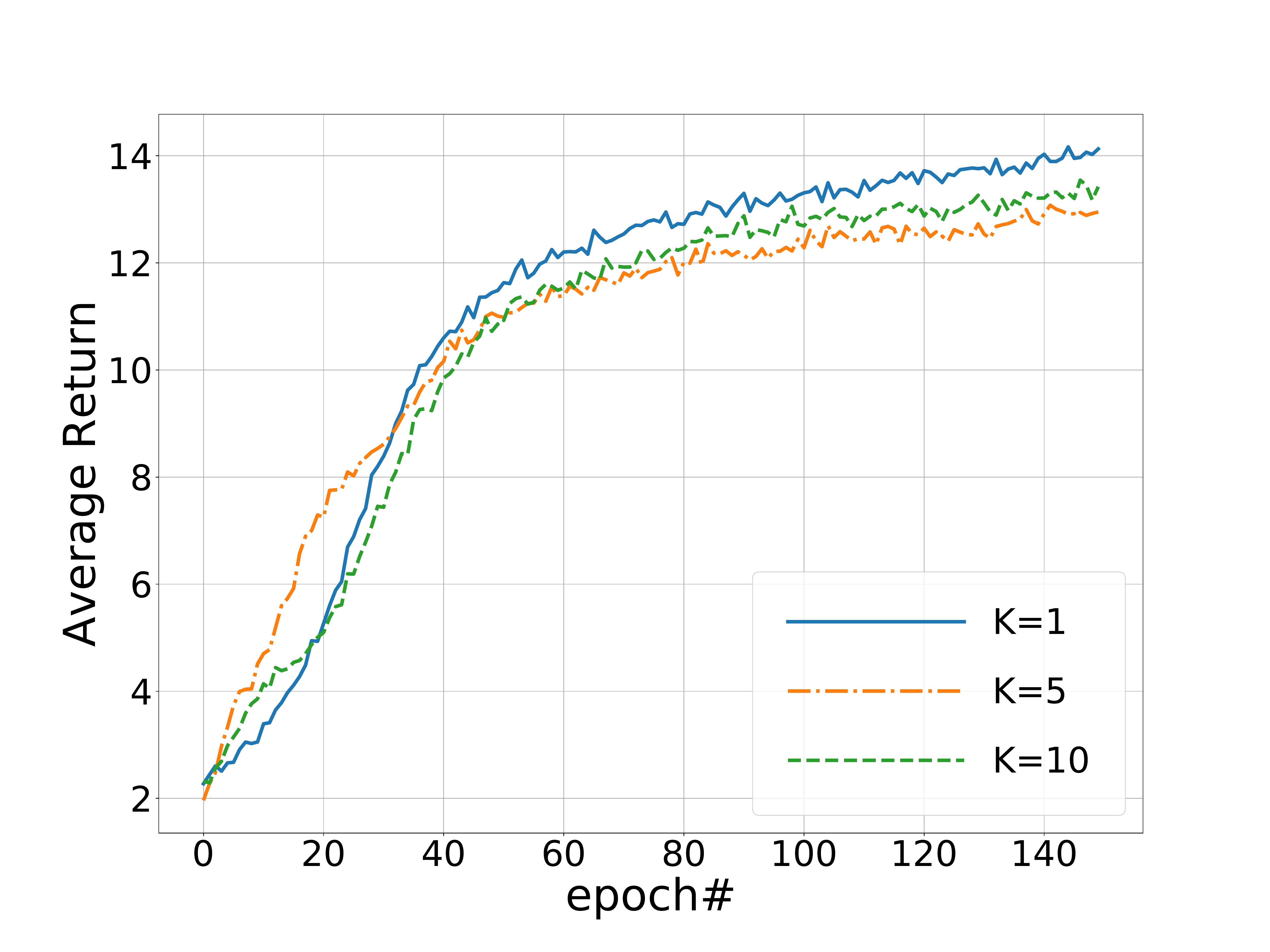}}
\subfigure[Average Cost (limit is 0.2)]{\label{fig:k2}\includegraphics[height=33mm]{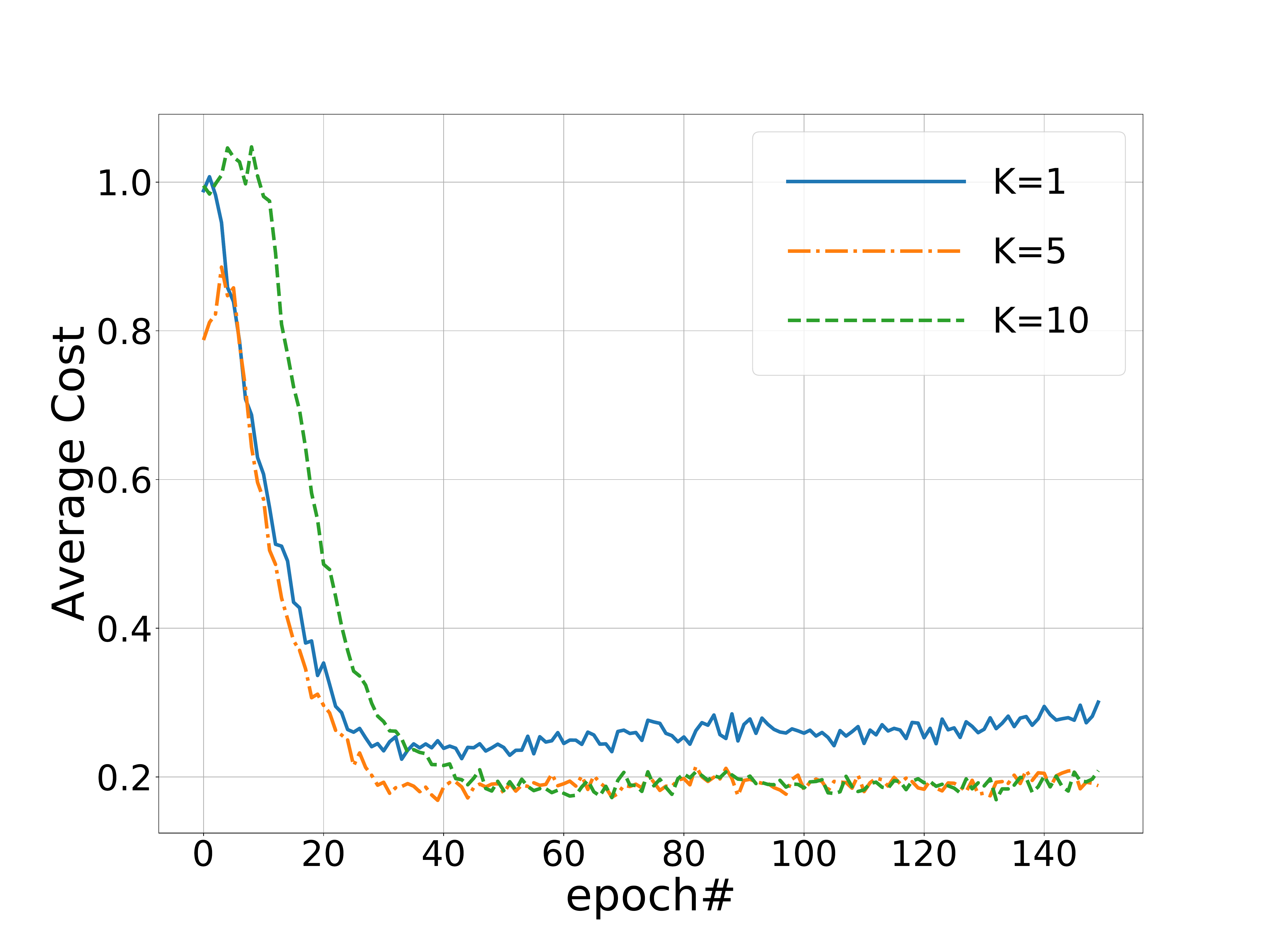}}
\subfigure[Dual Variable]{\label{fig:k3}\includegraphics[height=33mm]{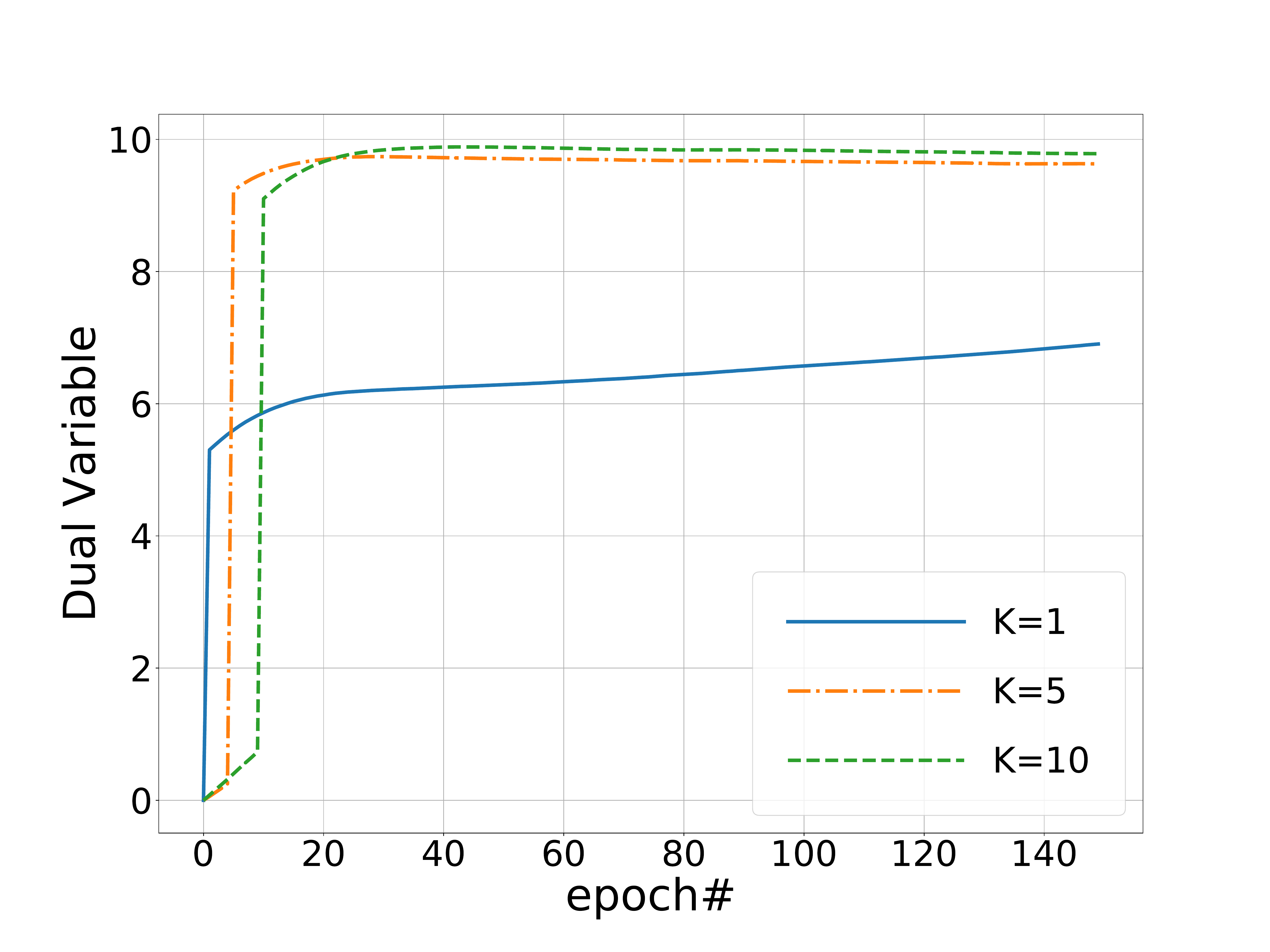}}
\caption{Effect of adjustment epoch $K^{\text{adj}}$.}
\label{fig:k}
\end{figure*}

\noindent \textbf{Effect of $K^{\text{adj}}$.} Figure \ref{fig:k} shows the effect of adjustment epoch $K^{\text{adj}}$ on the performance of APDO, where we experiment with $K^{\text{adj}}\in \{1, 5, 10\}$. It is observed that using a smaller $K^{\text{adj}}$ avoids slow incremental dual update earlier, but due to limited amount of available samples in the replay buffer the off-policy dual estimate $\lambda^{\text{OFF}}$ could be highly biased and inaccurate.  On the other hand, using a larger $K^{\text{adj}}$ provides a more accurate dual estimate at the expense of delayed adjustment.

\end{document}